\documentclass[10pt,twocolumn,letterpaper]{article}

\usepackage{multirow} 
\usepackage[pagenumbers]{cvpr} 
\usepackage{soul}
\usepackage[accsupp]{axessibility}  
\definecolor{cvprblue}{rgb}{0.21,0.49,0.74}
\usepackage[pagebackref,breaklinks,colorlinks,allcolors=cvprblue]{hyperref}

\def\paperID{39} 
\def\confName{CVPR}
\def\confYear{2026}

\title{From Frames to Events: Rethinking Evaluation in Human-Centric Video Anomaly Detection}

\author{
  Narges Rashvand \quad Shanle Yao  \quad Armin Danesh Pazho\\ 
  Babak Rahimi Ardabili \quad Hamed Tabkhi\\
  University of North Carolina at Charlotte\\
  Charlotte, NC, USA\\
  {\tt\small \{nrashvan,  syao, adaneshp, brahimia, htabkhiv\}@charlotte.edu}
}

\begin{document}
\maketitle
\begin{abstract}
Pose-based Video Anomaly Detection (VAD) has gained significant attention for its privacy-preserving nature and robustness to environmental variations. However, traditional frame-level evaluations treat video as a collection of isolated frames, fundamentally misaligned with how anomalies manifest and are acted upon in the real world. In operational surveillance systems, what matters is not the flagging of individual frames, but the reliable detection, localization, and reporting of a coherent anomalous
 \emph{event}— a contiguous temporal episode with an identifiable onset and duration. Frame-level metrics are blind to this distinction, and as a result, they systematically overestimate model performance for any deployment that requires actionable, event-level alerts. In this work, we propose a shift toward an event-centric perspective in VAD. We first audit widely used VAD benchmarks, including SHT~\cite{Liu_2018_CVPR}, CHAD~\cite{danesh2023chad}, NWPUC~\cite{cao2023new}, and HuVAD~\cite{pazho2025towards}, to characterize their event structure. We then introduce two strategies for temporal event localization: a score-refinement pipeline with hierarchical Gaussian smoothing and adaptive binarization, and an end-to-end Dual-Branch Model that directly generates event-level detections. Finally, we establish the first event-based evaluation standard for VAD by adapting Temporal Action Localization metrics, including $tIoU$-based event matching and multi-threshold $F_1$ evaluation. Our results quantify a substantial performance gap: while all SoTA models achieve frame-level AUC-ROC exceeding 52\% on the NWPUC~\cite{cao2023new}, their event-level localization precision falls below 10\% even at a minimal $tIoU=0.2$, with an average event-level $F_1$ of only 0.11 across all thresholds. The code base for this work is available at https://github.com/TeCSAR-UNCC/EventCentric-VAD.
\end{abstract}

\section{Introduction}
\label{sec:intro}
\begin{table}[htp]
\centering
\caption{Comparison between frame-level and event-level evaluation paradigms for video anomaly detection (VAD).}
\label{tab:metrics}
\resizebox{1\columnwidth}{!}{%
\begin{tabular}{l|ccccc}
\toprule[\heavyrulewidth] \midrule
\textbf{Aspects}     & \textbf{Frame-Level}  & \textbf{Event-Level}  \\ \midrule
Unit of analysis & Individual frames & Temporal events \\
Consideration of temporal continuity   & No & Yes  \\
Boundary localization  & Cannot measure &  Measured via tIoU \\
Operational trust & Low & High \\
Sensitivity to class imbalance & High & Lower \\
Suitable for real-world deployment & Limited & Yes \\

\midrule \bottomrule[\heavyrulewidth]
\end{tabular}%
}
\end{table}

Video Anomaly Detection (VAD) is a fundamental challenge in computer vision, essential for smart surveillance~\cite{fahrmann2024anomaly,mishra2024skeletal}, healthcare monitoring~\cite{ali2025anomaly,ghorbani2025examining,galvao2024anomaly}, and traffic analysis~\cite{khan2022anomaly,rahmanidehkordi2024traffic}. The objective of VAD is to automatically identify abnormal activities that deviate from expected behavior in video streams. In recent years, human-centric VAD has increasingly shifted from appearance-based representations toward pose-based methods~\cite{rashvand2025shopformer,noghre2024human}, which model the skeletal structure of human bodies, offering privacy-aware abstractions that are robust to illumination and background variations. Despite significant architectural advances for learning spatio-temporal patterns from pose sequences, these models are almost universally evaluated using frame-level metrics like AUC-ROC~\cite{hirschorn2023normalizing}, a protocol in which each frame is treated as an isolated snapshot and models are assessed on their ability to distinguish normal from anomalous frames across decision thresholds. We argue that this evaluation paradigm is fundamentally inadequate for real-world deployment, for four compounding reasons.

\textbf{Misalignment with the temporal nature of anomalies.}
Anomalies such as shoplifting, fighting, or falling are inherently continuous processes: they have a beginning, a progression, and an end. What makes them anomalous is not any single frame in isolation, but their trajectory across time. Evaluating models frame-by-frame reduces this temporal phenomenon to a series of independent classification decisions, measuring whether a frame is flagged rather than whether the underlying \emph{event}, a contiguous temporal episode with a discernible onset and duration, is detected and localized. This is a category error: the unit of measurement does not match the unit of the phenomenon.

\textbf{Disconnect from operational requirements.}
In deployed surveillance systems, operators do not respond to frames; they respond to incidents. A useful detection system must produce stable, coherent alerts corresponding to real behavioral episodes, not high-frequency frame-wise flags. Frame-level evaluation ignores this requirement entirely: a model that flickers between normal and anomalous predictions on consecutive frames may score well on AUC-ROC while generating a cascade of false alarms that render it operationally useless. Stakeholder studies confirm that law enforcement and security personnel have a low tolerance for false positives and require context-aware, temporally stable notifications to maintain trust in AI-assisted systems~\cite{ardabili2024exploring,ardabili2025co}. Temporal stability is therefore not a refinement; it is a prerequisite for societal acceptance.

\textbf{Performance overestimation by frame-level metrics}. A high AUC-ROC does not imply that a model can localize when an anomalous event begins or ends. Because frame-level metrics aggregate over all frames independently, a model can achieve strong AUC-ROC scores by correctly classifying the majority of normal frames while failing entirely to localize the anomalous episode within a sequence. This gives a misleading impression of deployment readiness. As we demonstrate empirically, all state-of-the-art models exceed 61\% AUC-ROC on the NWPUC benchmark~\cite{cao2023new}, yet their event-level precision falls below 10\% at $tIoU = 0.2$, a performance gap that frame-level evaluation conceals entirely.

\textbf{Event-aware model design}. The historical reliance on frame-level metrics has incentivized architectures optimized for frame-wise classification rather than temporal reasoning. Models that score well under the current paradigm are not necessarily learning to reason about event structure as they are learning to classify frames. Shifting to event-level evaluation does not merely refine measurement; it redirects the community toward architectures that model temporal coherence as a first-class objective.

Taken together, these motivations call for a fundamental shift from frame-level analysis to an event-centric evaluation paradigm in pose-based VAD. To this end, we propose a comprehensive framework with the following contributions:

\begin{itemize}
    \item We conduct the first systematic characterization of widely used VAD benchmarks,  SHT~\cite{Liu_2018_CVPR}, CHAD~\cite{danesh2023chad}, HuVAD~\cite{pazho2025towards}, and NWPUC~\cite{cao2023new}, from an event-centric perspective, analyzing their temporal event structure and suitability for event-level evaluation.

    \item We introduce a three-stage score-refinement pipeline; comprising hierarchical Gaussian smoothing, adaptive binarization, and physical constraint filtering, transforming frame-level anomaly scores into coherent temporal events by interpolating event boundaries from frame-level score sequences. Adapting Temporal Action Localization metrics~\cite{zhang2022actionformer,gokay2025skeleton}, we establish the first $tIoU$-based event matching and multi-threshold $F_1$ evaluation protocol for VAD, quantifying the true performance gap between frame-wise classification and event-level localization.

    \item We introduce an end-to-end Dual-Branch Model that employs multi-scale temporal windows to directly generate event-level detections, advancing the field beyond frame-wise classification toward fully automatic, temporally coherent event detection.
\end{itemize}
\section{Related Works}
\label{sec:related}

Traditional VAD has evolved from handcrafted-feature approaches to deep learning-based methods that learn normality from data. In modern semi-supervised VAD, most models are trained only on normal samples and identify anomalies through deviations in reconstruction, prediction, or distribution modeling \cite{markovitz2020gepc, jain2021posecvae, zeng2021hierarchical}. In pose-based VAD specifically, many methods follow a reconstruction-oriented paradigm in which the model learns regular human motion patterns and assigns higher anomaly scores to pose sequences that are difficult to reconstruct or predict \cite{ yu2023regularity, chen2023multiscale, huang2022hierarchical, hirschorn2023stgnf, rodrigues2020multi,noghre2024tsgad,noghre2025humancentricvideoanomalydetection}. Although these methods differ in architecture, they generally share the same assumption that anomaly detection can be reduced to frame-wise or short-window score estimation under a normality-learning objective.

Most existing VAD methods are evaluated using frame-level metrics such as AUC-ROC, where each frame is treated as an independent sample. This protocol has become the dominant benchmark standard because it provides a simple and unified way to compare models. However, it is increasingly clear that frame-level evaluation does not align well with real-world deployment, where operators care about coherent anomaly events, stable alerts, and actionable start-end localization rather than isolated anomalous frames\cite{doshi2021online,doshi2023towards, Doshi_2022_WACV}. Recent studies on real-world deployment of VAD systems have shown that the transition from controlled benchmarks to operational environments introduces challenges that are not captured by conventional offline frame-level evaluation, including unstable detections, threshold sensitivity, evolving environments, and the need for adaptation during deployment \cite{yao2025lab, yao2024evaluating, yao2025alfred, yao2026offline}. These findings suggest that high frame-level performance does not necessarily translate into reliable field performance, and that evaluation should better reflect operational utility.

This gap is even more pronounced from an event-centric perspective. Real anomalies such as shoplifting, falling, or fighting are temporally extended processes, yet frame-level scoring often fragments them into unstable predictions across adjacent frames. While related areas such as temporal action localization evaluate predictions as matched temporal intervals, event-level reasoning remains underexplored in VAD. Even recent attempts to use multimodal large language models for anomaly detection still commonly formulate the task as binary decision making over short video clips, such as 1-second segments, rather than explicit event localization over continuous time \cite{yao2026multimodal}. Therefore, despite growing interest in real-world VAD, the field still lacks a systematic event-level perspective that connects benchmark evaluation, temporal coherence, and deployment-oriented anomaly notification.

\section{Event-centric Characterization of VAD benchmarks}
\label{sec:dataset}

\begin{table}[htpb]
\centering
\caption{Frame-level and event-level statistics of VAD benchmarks.}
\label{tab:frame_event}
\resizebox{1\columnwidth}{!}{%
\begin{tabular}{ll|cccc}
\toprule
\textbf{Granularity} & \textbf{Characteristic} & \textbf{SHT\cite{Liu_2018_CVPR}}& \textbf{CHAD\cite{danesh2023chad}} & \textbf{HuVAD\cite{pazho2025towards}} & \textbf{NWPUC}\cite{cao2023new} \\
\midrule
\multirow{2}{*}{Frame} 
& Normal Frames & 24,077 & 67,303 & 694,415 & 318,793 \\
& Anomalous Frames & 16,714 & 59,172 & 225,075 & 65,266 \\

\midrule

\multirow{3}{*}{Event} 
& Anomalous Events & 121 & 190 & 1,691 & 137 \\
& Avg. Duration (f) & 138.13 & 311.43 & 133.10 & 476.39 \\

\bottomrule
\end{tabular}%
}
\end{table}

We establish a foundation by auditing existing VAD datasets, SHT \cite{Liu_2018_CVPR}, CHAD \cite{danesh2023chad}, HuVAD \cite{pazho2025towards}, and NWPUC \cite{cao2023new}, from an event-centric perspective. While traditional VAD relies on a frame-level binary masks $G=\{g_1, g_2, ..., g_n\}$, where $g_t=1$ denotes an anomalous frame and $g_t=0$ denotes a normal frame, this granularity often overlooks the semantic continuity of real-world anomalies. In VAD, an anomalous event is a continuous sequence of frames representing a semantically coherent activity, such as shoplifting, running, or fighting, occurring without interruption. To address this, we define an anomalous event $E_i$ as a contiguous subsequence $E_i=\{f_t, f_{t+1}, ..., f_{t+k}\}$ derived from the temporal transitions in $G$. Specifically, an event $E_i$ is initiated by the state shifts from 0 to 1 at $f_t$ and returns to 0 at $f_{t+k+1}$.

Our event-centric analysis reveals that the widely used SHT \cite{Liu_2018_CVPR} dataset contains micro-events, which are anomalous sequences spanning only a few frames. These likely represent manual annotation noise rather than semantically meaningful human actions. From a physical perspective, a human-centric anomaly cannot be meaningfully captured in a fraction of a second. Consequently, we audited the SHT \cite{Liu_2018_CVPR} test set by cross-referencing binary masks with the original videos, filtering out these physically impossible events to ensure every anomaly aligns with actual human movement dynamics. This cleaned version of SHT~\cite{Liu_2018_CVPR} forms the basis for all subsequent benchmarking and analysis in this work.

\begin{figure*}[]
    \centering
    \includegraphics[clip,trim={0 0 0 0},width=2\columnwidth]{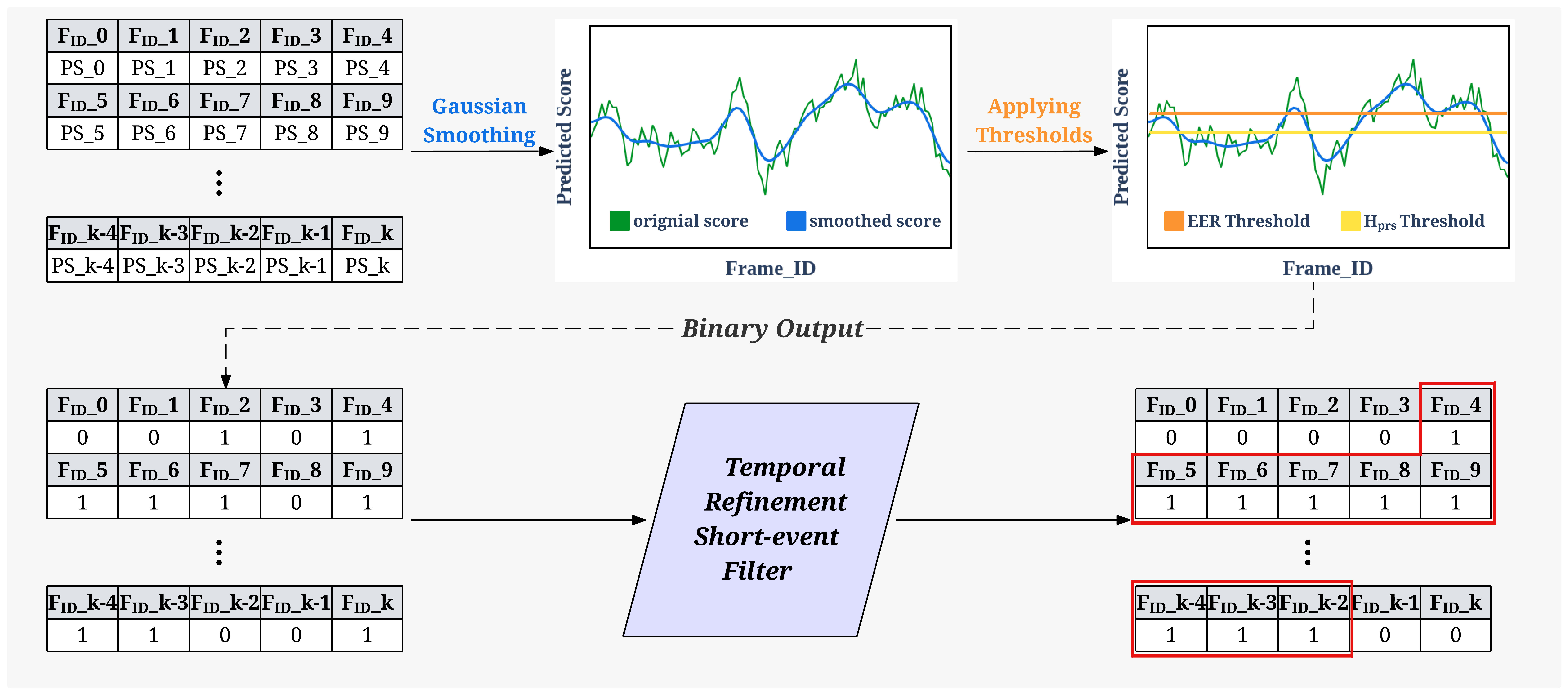}
    \caption{The proposed three-stage Frame-to-Event Transformation framework. Raw anomaly scores undergo hierarchical Gaussian smoothing to surpass high-frequency noise. Adaptive thresholds ($\tau_{EER}$ and $\tau_{H_{prs}}$) are then applied to the smoothed signal to generate a binary output. Finally, a temporal refinement and short-event filter resolve fragmented detections to produce semantically coherent anomalous events (red boxes) aligned with human motion dynamics.}
    \label{fig:manipulation_mode}
\end{figure*}

As shown in \cref{tab:frame_event}, the datasets vary significantly in scale and temporal characteristics, ranging from the relatively compact cleaned version of STH~\cite{Liu_2018_CVPR} to the massive HuVAD~\cite{pazho2025towards} dataset. HuVAD~\cite{pazho2025towards} contains the largest number of anomalous events, with 1,691 events in the test set. This provides a significantly broader distribution compared to CHAD \cite{danesh2023chad} (190 events), SHT~\cite{Liu_2018_CVPR} (121 events), and NWPUC~\cite{cao2023new} with 137 events. Regarding the average duration of anomalous events, the benchmarks represent a broad temporal spectrum. HuVAD~\cite{pazho2025towards} has the shortest events, averaging 133 frames, while NWPUC~\cite{cao2023new} contains the longest, averaging nearly 476 frames. CHAD~\cite{danesh2023chad} and SHT~\cite{Liu_2018_CVPR} fall between these two levels.


\section{Methodology}
\label{sec:methodology}

\begin{figure*}[]
    \centering
    \includegraphics[clip,trim={0 0 0 0},width=2\columnwidth]{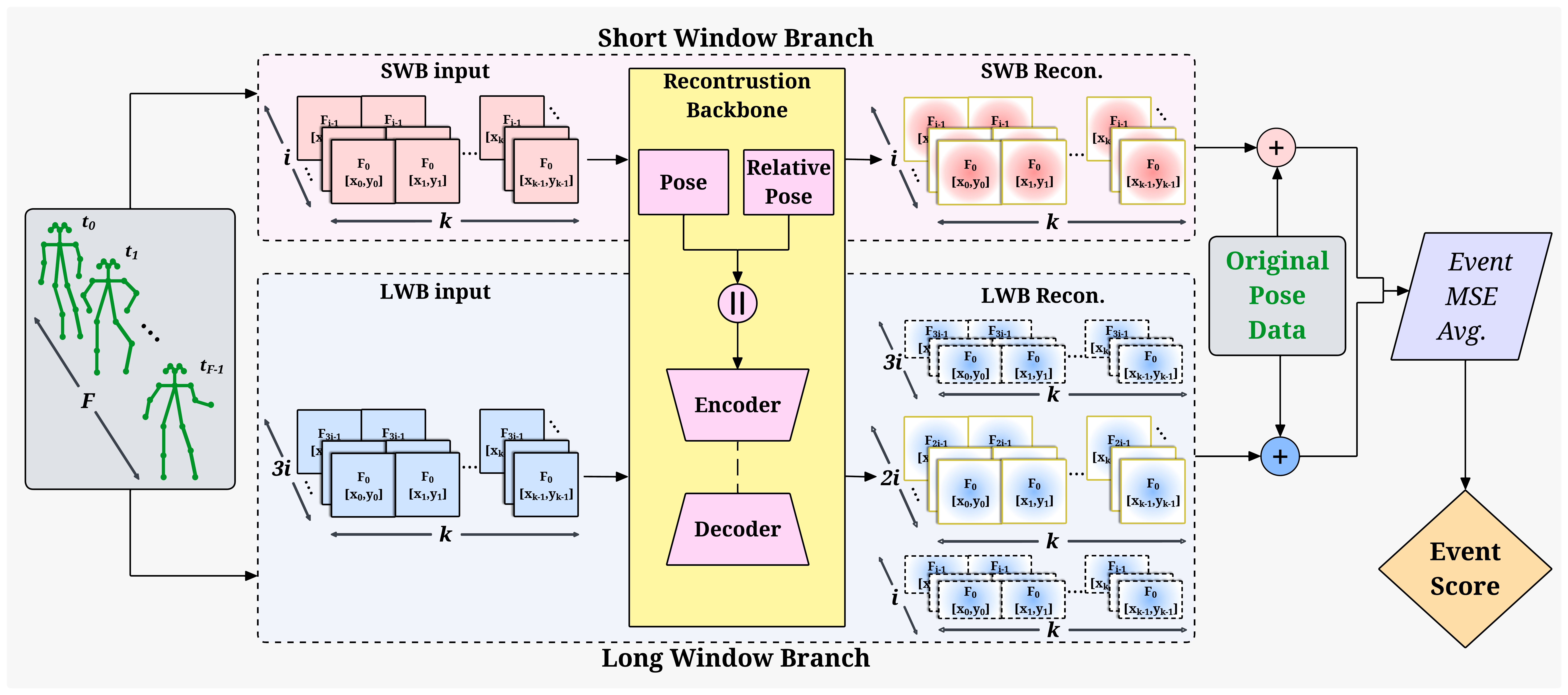}
    \caption{branch event-level anomaly detection framework. Given an input pose sequence, the model processes the data through two parallel branches: a Short Window Branch (SWB) with temporal length $i$ and a Long Window Branch (LWB) with temporal length $3i$. Both branches share the same transformer-based reconstruction backbone \cite{noghre2025humancentricvideoanomalydetection}, which jointly models absolute pose and relative pose through an encoder-decoder architecture. During inference, each branch produces frame-wise reconstruction errors. The center portion of the LWB error sequence is temporally aligned with the SWB target window, and the aligned frame-wise scores are fused to form a context-regularized anomaly response. The fused scores are then temporally pooled over the target window to produce a single event-level anomaly score.}
    \label{fig:model}
\end{figure*}



The proposed methodology is designed to shift the paradigm of VAD from traditional, isolated frame-level assessment to a semantically coherent event-level analysis. To achieve this, we introduce two distinct strategies for temporal localization. The first strategy involves a score-refinement pipeline (Manipulation Mode), which applies a three-stage pipeline to transform frame-level anomaly scores into discrete events. The second strategy introduces a Fully Automatic Method (Dual-Branch Reconstruction Event VAD), an architecture inherently designed for event detection, allowing for an end-to-end transition from raw pose data to actionable event-based alerts.

\subsection{Manipulation Mode: The Score-Refinement Pipeline}
\label{sec:f2e}
To bridge the gap between continuous frame-level scores and detection of discrete anomalous events, we propose a three-stage Frame-to-Event Transformation framework that interpolates event boundaries from frame-level anomaly score sequences. As illustrated in \cref {fig:manipulation_mode}, this framework comprises: (i) Hierarchical Gaussian smoothing, (ii) Multi-Threshold Binarization, and (iii) Temporal Event Extraction. This score-refinement pipeline is designed to suppress high-frequency noise, enforce temporal consistency, and filter out physically implausible detections to provide a more meaningful measure of real-world operational performance.
 
Raw anomaly scores $S$ often have significant high-frequency fluctuations, leading to fragmented and physically inconsistent detections. To mitigate this, we first implement a hierarchical Gaussian smoothing process. We iteratively apply 1D Gaussian kernel $G_{\sigma}(t) = \frac{1}{\sqrt{2\pi\sigma^2}} \exp\left(-\frac{t^2}{2\sigma^2}\right)$ with increasing standard deviations $\sigma \in \{1, \dots, \sigma_{max}\}$. This process effectively suppresses local noise while preserving the global trend of the anomaly score, ensuring a more stable input for the subsequent binarization and event extraction stages.

The smoothed continuous scores are then mapped to a binary state $B \in \{0, 1\}$ through a decision threshold $\tau$.  To assess model robustness across different operational requirements, we utilize two different thresholding strategies, where each threshold $\tau$ is uniquely derived for a specific model-dataset pair. Specifically, after the evaluation of a VAD architecture on a benchmark, we perform a grid search over the anomaly score distribution to identify the following operational boundaries:
\begin{itemize}
\item The Equal Error Rate threshold ($\tau_{EER}$), identified as the intersection of the False Acceptance Rate (FAR) and False Rejection Rate (FPR) on the ROC curve, representing a balanced sensitivity threshold.  
\item The $H_{prs}$ threshold  ($\tau_{H_{prs}}$) \cite{yao2026offline}, designed to simulate real-world security requirements where false alarms must be minimized. This is achieved by selecting a stricter threshold that prioritizes precision over recall.
\end{itemize}
While binarization identifies potential anomalies, frame-level predictions often suffer from noise and temporal fragmentation. This occurs when anomaly scores oscillate rapidly around the decision boundary, causing a single continuous event to be incorrectly perceived as a series of isolated, fragmented detections. To enforce temporal coherence, we apply a two-step refinement process consisting of majority voting and physical constraint filtering. For a video sequence of length $n$, a stride-based temporal window of size $W$ with stride $S$ is used to stabilize binarization. Within each window, the dominant state (Normal vs. Anomalous) is determined by majority voting over the $W$ frames, and the resulting decision is propagated to the corresponding output segment.
As a final refinement step, we enforce a minimum temporal footprint for all predicted events. Based on the physical constraints of human motion dynamics, an anomaly cannot occur within a fraction of a second. Consequently, we apply a short-event filter that discards any predicted sequence with a duration shorter than $D_{min}$ frames, ensuring outputs align with the natural dynamics of human movement. 

\subsection{Dual-Branch Reconstruction Event VAD}
\label{sec:dual}
To align anomaly scoring with the temporally continuous nature of real-world abnormal behavior, we propose a dual-branch event-level reconstruction framework that performs \emph{temporally aligned cross-scale score fusion}. Rather than assigning an anomaly score to each frame independently, the model first estimates frame-wise reconstruction errors under two different temporal receptive fields and then aggregates them into a single event-level score. This design encourages temporal coherence while reducing unstable local responses that commonly appear in conventional frame-level reconstruction-based VAD.

\begin{table*}[]
\centering
\caption{Frame-level evaluation performance of pose-based VAD models, STG-NF\cite{hirschorn2023stgnf}, SPARTA\cite{noghre2025humancentricvideoanomalydetection}, TS-GAD \cite{noghre2024tsgad}, across VAD benchmark datasets, SHT\cite{Liu_2018_CVPR}, CHAD\cite{danesh2023chad}, NWPUC \cite{cao2023new}, HuVAD\cite{pazho2025towards}. Thresholds $\tau_{EER}$, and  $\tau_{H_{prs}}$ are extracted per model-dataset pair to support subsequent event-level analysis.}
\label{tab:frame_results}
\resizebox{\textwidth}{!}{
\begin{tabular}{l|cccc|cccc|cccc}
\toprule
& \multicolumn{4}{c|}{\textbf{STG-NF\cite{hirschorn2023stgnf}}} & \multicolumn{4}{c|}{\textbf{SPARTA\cite{noghre2025humancentricvideoanomalydetection}}} & \multicolumn{4}{c}{\textbf{TS-GAD\cite{noghre2024tsgad}}} \\
\cmidrule(lr){2-5} \cmidrule(lr){6-9} \cmidrule(lr){10-13}
\textbf{Metric} & \textbf{SHT\cite{Liu_2018_CVPR}} & \textbf{CHAD\cite{danesh2023chad}} & \textbf{NWPUC\cite{cao2023new}} & \textbf{HuVAD\cite{pazho2025towards}} & \textbf{SHT\cite{Liu_2018_CVPR}} & \textbf{CHAD\cite{danesh2023chad}} & \textbf{NWPUC\cite{cao2023new}} & \textbf{HuVAD\cite{pazho2025towards}} & \textbf{SHT\cite{Liu_2018_CVPR}} & \textbf{CHAD\cite{danesh2023chad}} & \textbf{NWPUC\cite{cao2023new}} & \textbf{HuVAD\cite{pazho2025towards}} \\
\midrule
AUC-ROC & 0.866 & 0.570 & 0.617 & 0.520 & 0.861 & 0.569 & 0.635 & 0.649 & 0.807 & 0.553 & 0.618 & 0.623 \\
AUC-PR  & 0.838 & 0.569 & 0.276 & 0.252 & 0.831 & 0.517 & 0.258 & 0.327 & 0.740 & 0.499 & 0.300 & 0.305 \\
EER     & 0.218 & 0.458 & 0.408 & 0.478 & 0.225 & 0.444 & 0.408 & 0.409 & 0.256 & 0.461 & 0.418 & 0.417 \\
\midrule
$F1@{\tau_{EER}}$ & 0.746 & 0.52 & 0.330 & 0.347 & 0.738 & 0.538 & 0.329 & 0.413 & 0.703 & 0.521 & 0.322 & 0.406 \\
$F1@{\tau_{H_{prs}}}$ & 0.744 & 0.518 & 0.323 & 0.377 & 0.729 & 0.536 & 0.332 & 0.405 & 0.692 & 0.521 & 0.322 & 0.421 \\
\bottomrule
\end{tabular}
}
\end{table*}

\begin{table*}[]
\centering
\caption{Event-level performance of pose-based VAD models, including STG-NF\cite{hirschorn2023stgnf}, SPARTA\cite{noghre2025humancentricvideoanomalydetection}, TS-GAD\cite{noghre2024tsgad}, and our proposed framework, across the SHT\cite{Liu_2018_CVPR}, CHAD\cite{danesh2023chad}, NWPUC\cite{cao2023new}, and HuVAD\cite{pazho2025towards} datasets, using  $\tau_{EER}$. Results are compared across three configurations: Baselines (frame grouping), Baseline+Post-processing (using our refinement pipeline), and our natively event-aware model.}
\label{tab:eer_thr}
\resizebox{\textwidth}{!}{%
\begin{tabular}{llccccccccccccc}
\toprule
\toprule
& \multicolumn{1}{l|}{}               & \multicolumn{3}{c|}{\textbf{$tIOU=0.5$}}                                   & \multicolumn{3}{c|}{\textbf{$tIOU=0.4$}}                                   & \multicolumn{3}{c|}{\textbf{$tIOU=0.3$}}                                   & \multicolumn{3}{c|}{\textbf{$tIOU=0.2$}}                                   & \textbf{}                  \\ \cmidrule(lr){3-15} 
\textbf{Data}                                      & \multicolumn{1}{l|}{\textbf{Model}} & \textbf{Prec.} & \textbf{Rec.} & \multicolumn{1}{c|}{$\mathbf{F_1}$} & \textbf{Prec.} & \textbf{Rec.} & \multicolumn{1}{c|}{$\mathbf{F_1}$} & \textbf{Prec.} & \textbf{Rec.} & \multicolumn{1}{c|}{$\mathbf{F_1}$} & \textbf{Prec.} & \textbf{Rec.} & \multicolumn{1}{c|}{$\mathbf{F_1}$} & \textbf{Average $\mathbf{F_1}$}\\ \midrule
\multirow{10}{*}{\rotatebox{90}{SHT\cite{Liu_2018_CVPR}}}& \multicolumn{14}{c}{\textbf{Baselines}}      \\ \cmidrule(lr){2-15} 
& \multicolumn{1}{l|}{STG-NF\cite{hirschorn2023stgnf}}         & 50.59\%                & 71.07\%             & \multicolumn{1}{c|}{0.591}     & 60.00\%                & 84.30\%             & \multicolumn{1}{c|}{0.700}     & 61.76\%                & 86.78\%             & \multicolumn{1}{c|}{0.721}     & 64.71\%                & 90.91\%             & \multicolumn{1}{c|}{0.756}     & 0.692                        \\
& \multicolumn{1}{l|}{TS-GAD\cite{noghre2024tsgad}}         & 36.79\%                & 58.68\%             & \multicolumn{1}{c|}{0.452}              & 45.08\%                & 71.90\%             & \multicolumn{1}{c|}{0.554}              & 54.92\%                & 87.60\%             & \multicolumn{1}{c|}{0.675}              & 57.51\%                & 91.74\%             & \multicolumn{1}{c|}{0.707}              &0.597                        \\
 & \multicolumn{1}{l|}{SPARTA\cite{noghre2025humancentricvideoanomalydetection}}         & 44.62\%                & 68.60\%             & \multicolumn{1}{c|}{0.540}              & 52.15\%                & 80.17\%             & \multicolumn{1}{c|}{0.631}              & 56.45\%                & 86.78\%             & \multicolumn{1}{c|}{0.684}              & 0.596\%                & 91.74\%             & \multicolumn{1}{c|}{0.723}              & 0.644                        \\ \cmidrule(lr){2-15} 
 & \multicolumn{14}{c}{\textbf{Baselines + Proposed Post-Processing}}                                                                                                                                                                                                                                                                                                                                                            \\ \cmidrule(lr){2-15} 
& \multicolumn{1}{l|}{STG-NF\cite{hirschorn2023stgnf} w/ ES}   & 55.41\%            & 71.90\%         & \multicolumn{1}{c|}{0.625} & 63.06\%            & 81.82\%         & \multicolumn{1}{c|}{0.712} & 66.88\%            & 86.78\%         & \multicolumn{1}{c|}{0.755} & 69.43\%            & 90.08\%         & \multicolumn{1}{c|}{0.784} & 0.7193           \\
& \multicolumn{1}{l|}{TS-GAD\cite{noghre2024tsgad} w/ ES}   & 41.38\%            & 59.50\%         & \multicolumn{1}{c|}{0.488}          & 50.00\%            & 71.90\%         & \multicolumn{1}{c|}{0.589}          & 59.70\%            & 85.95\%         & \multicolumn{1}{c|}{0.705}          & 63.79\%            & 91.74\%         & \multicolumn{1}{c|}{0.752} & 0.6335                    \\
& \multicolumn{1}{l|}{SPARTA\cite{noghre2025humancentricvideoanomalydetection} w/ ES}   & 48.19\%            & 66.12\%         & \multicolumn{1}{c|}{0.557} & 56.63\%            & 77.69\%         & \multicolumn{1}{c|}{0.655} & 62.65\%            & 85.95\%         & \multicolumn{1}{c|}{0.724}          & 63.86\%            & 87.60\%         & \multicolumn{1}{c|}{0.738}          & 0.6685        \\ \cmidrule(lr){2-15} 
& \multicolumn{14}{c}{\textbf{Dual-Branch Event Detection}}                                                                                                                                                                                                       \\ \cmidrule(lr){2-15} 
& \multicolumn{1}{l|}{Dual (Ours)}    & 46.27\%            & 59.59\%         & \multicolumn{1}{c|}{0.521}          & 55.26\%            & 76.94\%         & \multicolumn{1}{c|}{0.643}          & 63.09\%            & 87.69\%         & \multicolumn{1}{c|}{0.734} & 66.45\%            & 87.12\%         & \multicolumn{1}{c|}{0.754}          & 0.662                    \\ \midrule

\multirow{10}{*}{\rotatebox{90}{CHAD\cite{danesh2023chad}}}  & \multicolumn{14}{c}{\textbf{Baselines}}     \\ \cmidrule(lr){2-15} 
& \multicolumn{1}{l|}{STG-NF\cite{hirschorn2023stgnf}}         & 7.31\%                & 23.68\%             & \multicolumn{1}{c|}{0.111}    & 11.69\%                & 37.89\%             & \multicolumn{1}{c|}{0.178}     & 17.05\%                & 55.26\%             & \multicolumn{1}{c|}{0.260}     & 22.08\%                & 71.58\%             & \multicolumn{1}{c|}{0.337}     & 0.222                        \\
 & \multicolumn{1}{l|}{TS-GAD\cite{noghre2024tsgad}}         & 3.92\%                & 13.68\%             & \multicolumn{1}{c|}{0.061}              & 8.14\%                & 28.42\%             & \multicolumn{1}{c|}{0.126}              & 13.57\%                & 47.37\%             & \multicolumn{1}{c|}{0.211}              & 18.40\%                & 64.21\%             & \multicolumn{1}{c|}{0.286}              & 0.171                        \\
  & \multicolumn{1}{l|}{SPARTA\cite{noghre2025humancentricvideoanomalydetection}}         & 10.00\%                & 25.26\%             & \multicolumn{1}{c|}{0.143}              & 14.58\%                & 36.84\%             & \multicolumn{1}{c|}{0.209}              & 21.25\%                & 53.68\%             & \multicolumn{1}{c|}{0.304}              & 26.25\%                & 66.32\%             & \multicolumn{1}{c|}{0.376}              & 0.258                        \\ \cmidrule(lr){2-15} 
& \multicolumn{14}{c}{\textbf{Baselines + Proposed Post-Processing}}                                                                                                                                                                                                                                                                                                                                                            \\ \cmidrule(lr){2-15} 
& \multicolumn{1}{l|}{STG-NF\cite{hirschorn2023stgnf} w/ ES}   & 8.21\%             & 24.21\%         & \multicolumn{1}{c|}{0.122}          & 11.96\%            & 35.26\%         & \multicolumn{1}{c|}{0.178}          & 18.39\%            & 54.21\%         & \multicolumn{1}{c|}{0.274}          & 23.57\%            & 69.47\%         & \multicolumn{1}{c|}{0.352}          & 0.231                     \\
& \multicolumn{1}{l|}{TS-GAD\cite{noghre2024tsgad} w/ ES}   & 5.38\%             & 16.32\%         & \multicolumn{1}{c|}{0.080}          & 10.24\%            & 31.05\%         & \multicolumn{1}{c|}{0.154}          & 16.32\%            & 49.47\%         & \multicolumn{1}{c|}{0.245}          & 21.53\%            & 65.26\%         & \multicolumn{1}{c|}{0.323}          & 0.200                     \\
 & \multicolumn{1}{l|}{SPARTA\cite{noghre2025humancentricvideoanomalydetection} w/ ES}   & 10.91\%            & 25.26\%         & \multicolumn{1}{c|}{0.152} & 16.14\%            & 37.37\%         & \multicolumn{1}{c|}{0.225} & 23.18\%            & 53.68\%         & \multicolumn{1}{c|}{0.323} & 28.41\%            & 65.79\%         & \multicolumn{1}{c|}{0.396} & 0.274         \\ \cmidrule(lr){2-15} 
 & \multicolumn{14}{c}{\textbf{Dual-Branch Event Detection}}                                                                                                            \\ \cmidrule(lr){2-15} 
 & \multicolumn{1}{l|}{Dual (Ours)}    & 16.71\%            & 30.53\%         & \multicolumn{1}{c|}{0.216} & 25.07\%            & 45.79\%         & \multicolumn{1}{c|}{0.324} & 34.87\%            & 63.68\%         & \multicolumn{1}{c|}{0.451} & 39.77\%            & 72.63\%         & \multicolumn{1}{c|}{0.514} & 0.376           \\ \midrule
\multirow{10}{*}{\rotatebox{90}{NWPUC\cite{cao2023new}}}  & \multicolumn{14}{c}{\textbf{Baselines}}                                                                                                                                                                                                                                                                                                                                                                                               \\ \cmidrule(lr){2-15} 
 & \multicolumn{1}{l|}{STG-NF\cite{hirschorn2023stgnf}}         & 3.92\%                & 23.36\%             & \multicolumn{1}{c|}{0.067}     & 5.63\%                & 33.58\%             & \multicolumn{1}{c|}{0.096}     & 6.98\%                & 41.61\%             & \multicolumn{1}{c|}{0.119}     & 9.55\%                & 56.93\%             & \multicolumn{1}{c|}{0.163}     & 0.111                        \\
& \multicolumn{1}{l|}{TS-GAD\cite{noghre2024tsgad}}         & 3.89\%                & 21.32\%             & \multicolumn{1}{c|}{0.065}              & 5.23\%                & 28.68\%             & \multicolumn{1}{c|}{0.088}              & 7.65\%                & 41.91\%             & \multicolumn{1}{c|}{0.129}              & 9.93\%                & 54.41\%             & \multicolumn{1}{c|}{0.168}              & 0.112                        \\
 & \multicolumn{1}{l|}{SPARTA\cite{noghre2025humancentricvideoanomalydetection}}         & 3.76\%                & 23.36\%             & \multicolumn{1}{c|}{0.064}              & 5.28\%                & 32.85\%             & \multicolumn{1}{c|}{0.091}              & 7.28\%                & 45.26\%             & \multicolumn{1}{c|}{0.125}              & 9.98\%                & 62.04\%             & \multicolumn{1}{c|}{0.171}              & 0.113                        \\ \cmidrule(lr){2-15} 
  & \multicolumn{14}{c}{\textbf{Baselines + Proposed Post-Processing}}                                                       \\ \cmidrule(lr){2-15} 
& \multicolumn{1}{l|}{STG-NF\cite{hirschorn2023stgnf} w/ ES}   & 4.39\%             & 24.09\%         & \multicolumn{1}{c|}{0.074} & 5.99\%             & 32.85\%         & \multicolumn{1}{c|}{0.101} & 7.86\%             & 43.07\%         & \multicolumn{1}{c|}{0.132}          & 10.12\%            & 55.47\%         & \multicolumn{1}{c|}{0.171}          & 0.119                     \\
& \multicolumn{1}{l|}{TS-GAD\cite{noghre2024tsgad} w/ ES}   & 4.26\%             & 21.32\%         & \multicolumn{1}{c|}{0.071}          & 5.73\%             & 28.68\%         & \multicolumn{1}{c|}{0.095} & 8.22\%             & 41.18\%         & \multicolumn{1}{c|}{0.137}          & 10.72\%            & 53.68\%         & \multicolumn{1}{c|}{0.178} & 0.120                     \\
& \multicolumn{1}{l|}{SPARTA\cite{noghre2025humancentricvideoanomalydetection} w/ ES}   & 4.13\%             & 23.36\%         & \multicolumn{1}{c|}{0.070}          & 5.68\%             & 32.12\%         & \multicolumn{1}{c|}{0.096}          & 7.62\%             & 43.07\%         & \multicolumn{1}{c|}{0.129} & 10.98\%            & 62.04\%         & \multicolumn{1}{c|}{0.186} & {0.120}         \\ \cmidrule(lr){2-15} 
 & \multicolumn{14}{c}{\textbf{Dual-Branch Event Detection}}                                                                  \\ \cmidrule(lr){2-15} 
   & \multicolumn{1}{l|}{Dual (Ours)}    & 4.00\%             & 21.90\%         & \multicolumn{1}{c|}{0.068} & 5.39\%             & 33.58\%         & \multicolumn{1}{c|}{0.093}          & 8.42\%             & 57.66\%         & \multicolumn{1}{c|}{0.147} & 9.50\%             & 64.96\%         & \multicolumn{1}{c|}{0.166}          & {0.118}            \\ \midrule
\multirow{10}{*}{\rotatebox{90}{HuVAD\cite{pazho2025towards}}} & \multicolumn{14}{c}{\textbf{Baselines}}                      \\ \cmidrule(lr){2-15} 
& \multicolumn{1}{l|}{STG-NF\cite{hirschorn2023stgnf}}         & 15.12\%                & 24.48\%             & \multicolumn{1}{c|}{0.186}     & 22.78\%                & 36.90\%             & \multicolumn{1}{c|}{0.281}     & 30.60\%                & 49.56\%             & \multicolumn{1}{c|}{0.378}     & 39.43\%                & 63.87\%             & \multicolumn{1}{c|}{0.487}     & 0.333                        \\
& \multicolumn{1}{l|}{TS-GAD\cite{noghre2024tsgad}}         & 13.83\%                & 26.91\%             & \multicolumn{1}{c|}{0.182}              & 20.45\%                & 39.80\%             & \multicolumn{1}{c|}{0.270}              & 28.14\%                & 54.76\%             & \multicolumn{1}{c|}{0.371}              & 34.55\%                & 67.24\%             & \multicolumn{1}{c|}{0.456}              & 0.320                       \\
& \multicolumn{1}{l|}{SPARTA\cite{noghre2025humancentricvideoanomalydetection}}         & 17.19\%                & 31.99\%             & \multicolumn{1}{c|}{0.223}              & 23.86\%                & 44.41\%             & \multicolumn{1}{c|}{0.310}              & 32.02\%                & 59.61\%             & \multicolumn{1}{c|}{0.416}              & 39.99\%                & 74.45\%             & \multicolumn{1}{c|}{0.520}              & 0.367                       \\ \cmidrule(lr){2-15} 
& \multicolumn{14}{c}{\textbf{Baselines + Proposed Post-Processing}}                                                       \\ \cmidrule(lr){2-15} 
& \multicolumn{1}{l|}{STG-NF\cite{hirschorn2023stgnf} w/ ES}   & 15.85\%            & 24.31\%         & \multicolumn{1}{c|}{0.191}          & 23.91\%            & 36.66\%         & \multicolumn{1}{c|}{0.289}          & 31.78\%            & 48.73\%         & \multicolumn{1}{c|}{0.384}          & 41.50\%            & 63.63\%         & \multicolumn{1}{c|}{0.502}          & 0.341                    \\
& \multicolumn{1}{l|}{TS-GAD\cite{noghre2024tsgad} w/ ES}   & 15.18\%            & 26.97\%         & \multicolumn{1}{c|}{0.194}          & 22.38\%            & 39.74\%         & \multicolumn{1}{c|}{0.286}          & 30.00\%            & 53.28\%         & \multicolumn{1}{c|}{0.383}          & 37.26\%            & 66.17\%         & \multicolumn{1}{c|}{0.476}          & 0.334                    \\
& \multicolumn{1}{l|}{SPARTA\cite{noghre2025humancentricvideoanomalydetection} w/ ES}   & 17.80\%            & 31.4\%         & \multicolumn{1}{c|}{0.227}& 24.71\%            & 43.58\%         & \multicolumn{1}{c|}{0.315} & 33.46\%            & 59.02\%         & \multicolumn{1}{c|}{0.427} & 41.80\%            & 73.74\%         & \multicolumn{1}{c|}{0.533} & 0.375        \\ \cmidrule(lr){2-15} 
& \multicolumn{14}{c}{\textbf{Dual-Branch Event Detection}}                                                       \\ \cmidrule(lr){2-15} 
& \multicolumn{1}{l|}{Dual (Ours)}    & 22.20\%            & 33.18\%         & \multicolumn{1}{c|}{0.266} & 30.30\%            & 45.29\%         & \multicolumn{1}{c|}{0.363} & 39.31\%            & 58.75\%         & \multicolumn{1}{c|}{0.471} & 47.41\%            & 70.86\%         & \multicolumn{1}{c|}{0.568} & {0.417}           \\ 
 \bottomrule
 \bottomrule
\end{tabular}%
}
\end{table*}

Given an input skeleton sequence, where each frame contains 17 human joints, the model forms two parallel inputs: a \textit{Short Window Branch} (SWB) and a \textit{Long Window Branch} (LWB), as illustrated in Fig.~\ref{fig:model}. The SWB operates on a target window of length $i$, while the LWB observes a broader temporal neighborhood of length $3i$ centered around the same target segment. Both branches are processed by the same shared reconstruction backbone. Importantly, this backbone is transformer-based and jointly models absolute pose and relative pose through a shared encoder-decoder architecture. This choice is particularly well motivated in our setting, since transformer-based reconstruction generally benefits from richer temporal context and broader input coverage. In other words, the LWB not only captures longer-range motion dependencies, but also provides a context-enriched view of the same target segment, which is later used to regularize the more locally sensitive SWB response.

During inference, each branch produces a sequence of frame-wise reconstruction errors. Since the LWB spans a larger temporal interval, only its center portion is retained so that it is temporally aligned with the SWB target window. Let $\mathbf{e}^{\text{S}} \in \mathbb{R}^{i}$ denote the SWB frame-wise error sequence, and let $\mathbf{e}^{\text{L}} \in \mathbb{R}^{3i}$ denote the LWB frame-wise error sequence. We extract the aligned center segment from the long branch as
\begin{equation}
\tilde{\mathbf{e}}^{\text{L}} = \mathbf{e}^{\text{L}}[i+1:2i].
\label{eq:lwb_align}
\end{equation}
This operation establishes frame-index correspondence between the two branches, ensuring that both error sequences describe the same target interval under different temporal contexts.

After alignment, we perform frame-wise fusion across scales. For each frame position within the target window, the final fused anomaly response is defined as the average of the short-window and aligned long-window errors:
\begin{equation}
\mathbf{e}^{\text{fuse}} = \frac{1}{2}\left(\mathbf{e}^{\text{S}} + \tilde{\mathbf{e}}^{\text{L}}\right).
\label{eq:frame_fusion}
\end{equation}
This step can be interpreted as a form of \emph{context-guided score regularization}. The SWB is naturally more sensitive to instantaneous pose irregularities, but it is also more vulnerable to abrupt spikes caused by pose noise, transient ambiguity, or short-lived reconstruction instability. By contrast, the aligned LWB response is derived from a broader temporal context and therefore provides a more context-stable estimate for the same frame positions. Their fusion yields a temporally aligned consensus score that preserves local anomaly evidence while damping isolated short-window fluctuations.

Finally, the event-level anomaly score is obtained by temporally pooling the fused frame-wise responses over the target window:
\begin{equation}
E_{\text{event}} = \frac{1}{i}\sum_{t=1}^{i} e^{\text{fuse}}_t.
\label{eq:event_score}
\end{equation}
This final aggregation converts the fused frame-wise anomaly trajectory into a single score for the entire segment. As a result, the model does not treat the event as a collection of disconnected frame decisions, but instead scores it as a coherent temporal unit. Overall, the proposed scoring strategy consists of two coupled stages: \emph{cross-scale frame alignment and fusion}, followed by \emph{event-level temporal pooling}. This formulation is more expressive than directly averaging two branch-level scores, because it first enforces temporal correspondence at the frame level and only then aggregates the resulting consensus response into an event score. In this way, the final anomaly estimate reflects both local reconstruction difficulty and broader temporal consistency. The resulting prediction is therefore more stable, less fragmented, and better suited for event-level thresholding and temporally coherent anomaly localization in real-world surveillance scenarios.

\section{Experiments and Results}
\label{sec:results}

\begin{table*}[]
\centering
\caption{Event-level performance of pose-based VAD models, including STG-NF\cite{hirschorn2023stgnf}, SPARTA\cite{noghre2025humancentricvideoanomalydetection}, TS-GAD\cite{noghre2024tsgad}, and our proposed framework, across the SHT\cite{Liu_2018_CVPR}, CHAD\cite{danesh2023chad}, NWPUC\cite{cao2023new}, and HuVAD\cite{pazho2025towards} datasets, using $(\tau_{H_{prs}})$. Results are compared across three configurations: Baselines (frame grouping), Baseline+Post-processing (using our refinement pipeline), and our natively event-aware model.}
\label{tab:hprs_thr}
\resizebox{\textwidth}{!}{%
\begin{tabular}{llccccccccccccc}
\toprule
\toprule
& \multicolumn{1}{l|}{}               & \multicolumn{3}{c|}{\textbf{$tIOU=0.5$}}                                   & \multicolumn{3}{c|}{\textbf{$tIOU=0.4$}}                                   & \multicolumn{3}{c|}{\textbf{$tIOU=0.3$}}                                   & \multicolumn{3}{c|}{\textbf{$tIOU=0.2$}}                                   & \textbf{}                  \\ \cmidrule(lr){3-15} 
\textbf{Data}                                              & \multicolumn{1}{l|}{\textbf{Model}} & \textbf{Precision} & \textbf{Rec.} & \multicolumn{1}{c|}{$\mathbf{F_1}$} & \textbf{Precision} & \textbf{Rec.} & \multicolumn{1}{c|}{$\mathbf{F_1}$} & \textbf{Precision} & \textbf{Rec.} & \multicolumn{1}{c|}{$\mathbf{F_1}$} & \textbf{Precision} & \textbf{Rec.} & \multicolumn{1}{c|}{$\mathbf{F_1}$} & \textbf{Average $\mathbf{F_1}$} \\ \midrule
\multirow{10}{*}{\rotatebox{90}{SHT\cite{Liu_2018_CVPR}}}& \multicolumn{14}{c}{\textbf{Baselines}}                                                                                           \\ \cmidrule(lr){2-15} 
& \multicolumn{1}{l|}{STG-NF\cite{hirschorn2023stgnf}}         & 51.79\%                & 71.90\%             & \multicolumn{1}{c|}{0.602}    & 57.14\%                & 79.34\%             & \multicolumn{1}{c|}{0.664}     & 58.93\%                & 81.82\%             & \multicolumn{1}{c|}{0.685}     & 63.69\%                & 88.43\%             & \multicolumn{1}{c|}{0.740}     & 0.673                     \\
& \multicolumn{1}{l|}{TS-GAD\cite{noghre2024tsgad}}         & 41.38\%                & 59.50\%             & \multicolumn{1}{c|}{0.488}              & 50.00\%                & 71.90\%             & \multicolumn{1}{c|}{0.589}              & 57.47\%                & 82.64\%             & \multicolumn{1}{c|}{0.678}              & 62.07\%                & 89.26\%             & \multicolumn{1}{c|}{0.732}              & 0.622                        \\
& \multicolumn{1}{l|}{SPARTA\cite{noghre2025humancentricvideoanomalydetection}}         & 45.83\%                & 63.64\%             & \multicolumn{1}{c|}{0.532}              & 54.76\%                & 76.03\%             & \multicolumn{1}{c|}{0.636}              & 57.74\%                & 80.17\%             & \multicolumn{1}{c|}{0.671}              & 64.29\%                & 89.26\%             & \multicolumn{1}{c|}{0.747}              & 0.647                        \\ \cmidrule(lr){2-15} 
& \multicolumn{14}{c}{\textbf{Baselines + Proposed Post-Processing}}                                                \\ \cmidrule(lr){2-15} 
& \multicolumn{1}{l|}{STG-NF\cite{hirschorn2023stgnf} w/ ES}   & 56.58\%                & 71.07\%             & \multicolumn{1}{c|}{0.630}     & 62.50\%                & 78.51\%             & \multicolumn{1}{c|}{0.696}     & 65.79\%                & 82.64\%             & \multicolumn{1}{c|}{0.732}     &70.39\%                & 88.43\%             & \multicolumn{1}{c|}{0.783}     & 0.710                        \\
& \multicolumn{1}{l|}{TS-GAD\cite{noghre2024tsgad} w/ ES}   & 41.46\%                & 56.20\%             & \multicolumn{1}{c|}{0.477}              & 53.05\%                & 71.90\%             & \multicolumn{1}{c|}{0.610}              & 59.15\%                & 80.17\%             & \multicolumn{1}{c|}{0.680}              & 64.02\%                & 86.78\%             & \multicolumn{1}{c|}{0.736}              & 0.626                       \\
& \multicolumn{1}{l|}{SPARTA\cite{noghre2025humancentricvideoanomalydetection} w/ ES}   & 47.80\%                & 62.81\%             & \multicolumn{1}{c|}{0.542}              & 56.60\%                & 74.38\%             & \multicolumn{1}{c|}{0.642}              & 61.01\%                & 80.17\%             & \multicolumn{1}{c|}{0.692}              & 67.30\%                & 88.43\%             & \multicolumn{1}{c|}{0.764}              & 0.660                        \\ \cmidrule(lr){2-15} 
& \multicolumn{14}{c}{\textbf{Dual-Branch Event Detection}}                                                   \\ \cmidrule(lr){2-15} 
& \multicolumn{1}{l|}{Dual (Ours)}    & 44.44\%            & 59.59\%         & \multicolumn{1}{c|}{0.509}          & 56.30\%            & 72.81\%         & \multicolumn{1}{c|}{0.643}          & 64.83\%            & 87.69\%         & \multicolumn{1}{c|}{0.745}          & 69.63\%            & 86.69\%         & \multicolumn{1}{c|}{0.772}          & 0.667                    \\ \midrule
\multirow{10}{*}{\rotatebox{90}{CHAD\cite{danesh2023chad}}}  & \multicolumn{14}{c}{\textbf{Baselines}}                                                                                             \\ \cmidrule(lr){2-15} 
& \multicolumn{1}{l|}{STG-NF\cite{hirschorn2023stgnf}}         & 6.91\%                & 22.63\%             & \multicolumn{1}{c|}{0.105}     & 10.29\%                & 33.68\%             & \multicolumn{1}{c|}{0.157}     & 15.92\%                & 52.11\%             & \multicolumn{1}{c|}{0.243}     & 21.22\%                & 69.47\%             & \multicolumn{1}{c|}{0.325}     & 0.208                        \\
& \multicolumn{1}{l|}{TS-GAD\cite{noghre2024tsgad}}         & 3.92\%                & 13.68\%             & \multicolumn{1}{c|}{0.061}              & 8.14\%                & 28.42\%             & \multicolumn{1}{c|}{0.126}              & 13.57\%                & 47.37\%             & \multicolumn{1}{c|}{0.211}              & 18.4\%                & 64.21\%             & \multicolumn{1}{c|}{0.286}              & 0.171                       \\
& \multicolumn{1}{l|}{SPARTA\cite{noghre2025humancentricvideoanomalydetection}}         & 10.11\%                & 25.26\%             & \multicolumn{1}{c|}{0.144}              & 14.53\%                & 36.32\%             & \multicolumn{1}{c|}{0.207}              & 21.05\%                & 52.63\%             & \multicolumn{1}{c|}{0.300}              & 26.53\%                & 66.32\%             & \multicolumn{1}{c|}{0.378}              & 0.257                        \\ \cmidrule(lr){2-15} 
& \multicolumn{14}{c}{\textbf{Baselines + Proposed Post-Processing}}                                               \\ \cmidrule(lr){2-15} 
& \multicolumn{1}{l|}{STG-NF\cite{hirschorn2023stgnf} w/ ES}   & 7.75\%                & 23.16\%             & \multicolumn{1}{c|}{0.116}     & 11.62\%                & 34.74\%             & \multicolumn{1}{c|}{0.174}     & 17.25\%                & 51.58\%             & \multicolumn{1}{c|}{0.258}     & 23.42\%                & 70.00\%             & \multicolumn{1}{c|}{0.350}     & 0.224                        \\
 & \multicolumn{1}{l|}{TS-GAD\cite{noghre2024tsgad} w/ ES}   & 5.37\%                & 16.32\%             & \multicolumn{1}{c|}{0.080}              &10.23\%                & 31.05\%             & \multicolumn{1}{c|}{0.153}              & 16.29\%                & 49.47\%             & \multicolumn{1}{c|}{0.245}              & 21.49\%                & 65.26\%             & \multicolumn{1}{c|}{0.323}              & 0.200                        \\
& \multicolumn{1}{l|}{SPARTA\cite{noghre2025humancentricvideoanomalydetection} w/ ES}   & 10.61\%                & 24.74\%             & \multicolumn{1}{c|}{0.148}              & 16.03\%                & 37.37\%             & \multicolumn{1}{c|}{0.224}              & 22.57\%                & 52.63\%             & \multicolumn{1}{c|}{0.316}              & 28.22\%                & 65.79\%             & \multicolumn{1}{c|}{0.394}              & 0.270                        \\ \cmidrule(lr){2-15} 
& \multicolumn{14}{c}{\textbf{Dual-Branch Event Detection}}                                                    \\ \cmidrule(lr){2-15} 
& \multicolumn{1}{l|}{Dual (Ours)}    & 17.29\%            & 31.58\%         & \multicolumn{1}{c|}{0.223}          & 24.78\%            & 45.26\%         & \multicolumn{1}{c|}{0.320}          & 34.87\%            & 63.68\%         & \multicolumn{1}{c|}{0.451}          & 40.06\%            & 74.16\%         & \multicolumn{1}{c|}{0.520}          & 0.378                    \\ \midrule
\multirow{10}{*}{\rotatebox{90}{NWPUC\cite{cao2023new}}}  & \multicolumn{14}{c}{\textbf{Baselines}}                                                                                              \\ \cmidrule(lr){2-15} 
& \multicolumn{1}{l|}{STG-NF\cite{hirschorn2023stgnf}}         & 1.89\%                & 11.68\%             & \multicolumn{1}{c|}{0.032}     & 3.31\%                & 20.44\%             & \multicolumn{1}{c|}{0.057}     & 5.21\%                & 32.12\%             & \multicolumn{1}{c|}{0.089}     & 7.34\%                & 45.26\%             & \multicolumn{1}{c|}{0.126}     & 0.076                        \\
& \multicolumn{1}{l|}{TS-GAD\cite{noghre2024tsgad}}         & 2.30\%                & 13.24\%             & \multicolumn{1}{c|}{0.039}              & 4.10\%                & 23.53\%             & \multicolumn{1}{c|}{0.069}              & 5.38\%                & 30.88\%             & \multicolumn{1}{c|}{0.091}              & 6.79\%                & 38.97\%             & \multicolumn{1}{c|}{0.115}              & 0.070                        \\
& \multicolumn{1}{l|}{SPARTA\cite{noghre2025humancentricvideoanomalydetection}}         & 3.15\%                & 18.98\%             & \multicolumn{1}{c|}{0.054}              & 4.00\%                & 24.09\%             & \multicolumn{1}{c|}{0.068}              & 6.06\%                & 36.50\%             & \multicolumn{1}{c|}{0.104}              & 9.33\%                & 56.20\%             & \multicolumn{1}{c|}{0.160}              & 0.096                        \\ \cmidrule(lr){2-15} 
& \multicolumn{14}{c}{\textbf{Baselines + Proposed Post-Processing}}                                                \\ \cmidrule(lr){2-15} 
& \multicolumn{1}{l|}{STG-NF\cite{hirschorn2023stgnf} w/ ES}   & 2.26\%                & 12.41\%             & \multicolumn{1}{c|}{0.038}     & 3.86\%                & 21.17\%             & \multicolumn{1}{c|}{0.065}     & 5.85\%                & 32.12\%             & \multicolumn{1}{c|}{0.099}     & 7.85\%                & 43.07\%             & \multicolumn{1}{c|}{0.132}     & 0.083                       \\
& \multicolumn{1}{l|}{TS-GAD\cite{noghre2024tsgad} w/ ES}   & 2.73\%                & 13.97\%             & \multicolumn{1}{c|}{0.045}              & 4.45\%                & 22.79\%             & \multicolumn{1}{c|}{0.074}              & 5.88\%                & 30.15\%             & \multicolumn{1}{c|}{0.098}              & 7.32\%                & 37.50\%             & \multicolumn{1}{c|}{0.122}              & 0.085                       \\
& \multicolumn{1}{l|}{SPARTA\cite{noghre2025humancentricvideoanomalydetection} w/ ES}   & 3.33\%                & 18.25\%             & \multicolumn{1}{c|}{0.056}              & 4.67\%                & 25.55\%             & \multicolumn{1}{c|}{0.078}              & 6.53\%                & 35.77\%             & \multicolumn{1}{c|}{0.110}              & 10.27\%                & 56.20\%             & \multicolumn{1}{c|}{0.173}     & 0.104                        \\ \cmidrule(lr){2-15} 
 & \multicolumn{14}{c}{\textbf{Dual-Branch Event Detection}}                                                                                                             \\ \cmidrule(lr){2-15} 
& \multicolumn{1}{l|}{Dual (Ours)}    & 4.15\%             & 20.25\%         & \multicolumn{1}{c|}{0.069}          & 6.02\%             & 34.09\%         & \multicolumn{1}{c|}{0.102}          & 9.28\%             & 50.85\%         & \multicolumn{1}{c|}{0.157}          & 9.49\%             & 59.64\%         & \multicolumn{1}{c|}{0.164}          & 0.123                     \\ \midrule
\multirow{10}{*}{\rotatebox{90}{HuVAD\cite{pazho2025towards}}} & \multicolumn{14}{c}{\textbf{Baselines}}                                                                                                                                               \\ \cmidrule(lr){2-15} 
& \multicolumn{1}{l|}{STG-NF\cite{hirschorn2023stgnf}}         & 17.87\%                & 28.62\%             & \multicolumn{1}{c|}{0.220}     & 26.54\%                & 42.52\%             & \multicolumn{1}{c|}{0.326}     & 35.66\%                & 57.13\%             & \multicolumn{1}{c|}{0.439}     & 44.26\%                & 70.90\%             & \multicolumn{1}{c|}{0.545}     & 0.382                        \\
& \multicolumn{1}{l|}{TS-GAD\cite{noghre2024tsgad}}         & 15.35\%                & 30.10\%             & \multicolumn{1}{c|}{0.203}              & 21.95\%                & 43.05\%             & \multicolumn{1}{c|}{0.290}              & 30.85\%                & 60.50\%             & \multicolumn{1}{c|}{0.408}              & 37.24\%                & 73.03\%             & \multicolumn{1}{c|}{0.493}              & 0.349                        \\
& \multicolumn{1}{l|}{SPARTA\cite{noghre2025humancentricvideoanomalydetection}}         & 16.07\%                & 29.63\%             & \multicolumn{1}{c|}{0.208}              & 22.91\%                & 42.22\%             & \multicolumn{1}{c|}{0.297}              & 30.83\%                & 56.83\%             & \multicolumn{1}{c|}{0.399}              & 38.34\%                & 70.67\%             & \multicolumn{1}{c|}{0.497}              & 0.350                        \\ \cmidrule(lr){2-15} 
& \multicolumn{14}{c}{\textbf{Baselines + Proposed Post-Processing}}                                                                                                        \\ \cmidrule(lr){2-15} 
& \multicolumn{1}{l|}{STG-NF\cite{hirschorn2023stgnf} w/ ES}   & 18.46\%                & 28.44\%             & \multicolumn{1}{c|}{0.223}     & 27.52\%                & 42.40\%             & \multicolumn{1}{c|}{0.333}     & 36.66\%                & 56.48\%             & \multicolumn{1}{c|}{0.444}    & 45.83\%                & 70.61\%             & \multicolumn{1}{c|}{0.555}     & 0.389                       \\
& \multicolumn{1}{l|}{TS-GAD\cite{noghre2024tsgad} w/ ES}   & 16.80\%                & 30.28\%             & \multicolumn{1}{c|}{0.216}              & 24.02\%                & 43.29\%             & \multicolumn{1}{c|}{0.309}              & 33.02\%                & 59.49\%             & \multicolumn{1}{c|}{0.424}              & 39.88\%                & 71.85\%             & \multicolumn{1}{c|}{0.512}              & 0.365                        \\
& \multicolumn{1}{l|}{SPARTA\cite{noghre2025humancentricvideoanomalydetection} w/ ES}   & 17.14\%                & 29.45\%             & \multicolumn{1}{c|}{0.216}              & 23.85\%                & 40.98\%             & \multicolumn{1}{c|}{0.301}              & 32.31\%                & 55.53\%             & \multicolumn{1}{c|}{0.408}              & 40.50\%                & 69.60\%             & \multicolumn{1}{c|}{0.512}              & 0.359                        \\ \cmidrule(lr){2-15} 
& \multicolumn{14}{c}{\textbf{Dual-Branch Event Detection}}                                                                                                                                                                        \\ \cmidrule(lr){2-15} 
& \multicolumn{1}{l|}{Dual (Ours)}    & 22.77\%            & 33.97\%         & \multicolumn{1}{c|}{0.273}          & 30.29\%            & 46.48\%         & \multicolumn{1}{c|}{0.367}          & 39.00\%            & 58.27\%         & \multicolumn{1}{c|}{0.467}          & 48.25\%            & 69.30\%         & \multicolumn{1}{c|}{0.569}          & 0.419                    \\ 
\bottomrule
\bottomrule
\end{tabular}%
}
\end{table*}

\cref{tab:frame_results} presents the frame-level performance of three state-of-the-art pose-based VAD methods on four benchmark datasets, evaluated using standard ranking-based metrics, including AUC-ROC, AUC-PR, and EER. Based on these frame-level score distributions, we further derive two operating thresholds, $\tau_{EER}$ and $\tau_{H_{prs}}$, which are later used for event-level analysis. To better reflect practical decision-making behavior, we also report frame-level $F_1$ scores at $F_1@\tau_{EER}$ and $F_1@\tau_{H_{prs}}$, by binarizing each frame as either normal or anomalous. As shown in \cref{tab:frame_results}, several methods achieve relatively strong threshold-independent ranking performance, particularly in terms of AUC-ROC, while also exhibiting similar EER values. However, such metrics may overestimate practical detection capability, as they primarily measure score separability rather than the precision of final boundary predictions. Once a fixed threshold is applied, the evaluation becomes more clearer, and the resulting $F_1$ scores provide a more realistic assessment of deployment-oriented performance. This effect is especially evident on the more challenging datasets, particularly NWPUC\cite{cao2023new} and HuVAD\cite{pazho2025towards}, revealing an unstable precision-recall trade-off under binary decision settings.

\cref{tab:eer_thr} and \cref{tab:hprs_thr} report event-level anomaly detection results under two different thresholding strategies, namely $\tau_{EER}$ and $\tau_{H_{prs}}$, respectively. In both cases, we evaluate three settings that represent different ways of transferring frame-level VAD outputs to event-level detection. 

In the Baseline setting, the binary frame-level decision masks obtained from \cref{tab:frame_results} are directly converted into anomaly events and evaluated at different tIoU thresholds. Across both thresholding strategies, this direct transfer consistently yields limited event-level performance, especially on the more challenging datasets. Although some frame-level operating points remain acceptable, the resulting event-level $F_1$ scores are often substantially lower, indicating that frame-wise binary predictions are highly vulnerable to fragmented activations, unstable temporal boundaries, and short spurious detections once evaluated as complete events rather than isolated frames. 

Applying the proposed event smoothing strategy from \cref{sec:f2e} generally improves event-level $F_1$ under both $\tau_{EER}$ and $\tau_{H_{prs}}$, confirming that suppressing unreasonable short segments helps bridge part of the gap between frame-level decisions and coherent event localization. This effect is particularly clear on SHT\cite{Liu_2018_CVPR}, where post-processing consistently boosts the event-level results across multiple tIoU settings, while on CHAD\cite{danesh2023chad}, NWPUC\cite{cao2023new}, and HuVAD\cite{pazho2025towards} the gains are present but more modest, suggesting that heuristic smoothing alone cannot fully address the event fragmentation problem on more difficult datasets. 

In contrast, the proposed Dual-Branch Event Detection approach in \cref{sec:dual} offers a fundamentally different transition from frame-level VAD to event-level reasoning by directly modeling anomaly events instead of relying on post process correction. Importantly, this advantage remains consistent under both thresholding schemes: the dual-branch design delivers the strongest and most stable improvements on CHAD\cite{danesh2023chad} and HuVAD\cite{pazho2025towards}, where direct threshold transfer and simple smoothing are insufficient, demonstrating that explicit event-level modeling is more effective for temporally ambiguous and irregular anomaly patterns. On SHT\cite{Liu_2018_CVPR}, however, the post-processing strategy remains highly competitive and often outperforms the dual-branch design, implying that when frame-level predictions are already temporally clean, lightweight smoothing may be sufficient. For NWPUC\cite{cao2023new}, all methods remain relatively weak under both thresholds, further showing that converting frame-level anomaly scores into reliable event-level detections is intrinsically difficult on highly challenging data. Our results reveal a substantial gap between conventional frame-level evaluation and practically meaningful event-level performance, indicating that existing VAD methods are considerably less effective than traditional metrics suggest. For example, although state-of-the-art models achieve frame-level AUC-ROC scores above 61\% on NWPUC~\cite{cao2023new}, their event-level localization performance falls below 10\% at $tIoU=0.5$. This discrepancy highlights a fundamental limitation of frame-based evaluation: while it measures score separability, it does not reflect whether anomalies can be localized as coherent temporal events. In contrast, our event-level framework provides a more realistic benchmark for surveillance scenarios, where the goal is to detect actionable anomaly events rather than isolated abnormal frames.
Importantly, this observation holds across different thresholding strategies. As shown in the two tables, switching from $\tau_{EER}$ to $\tau_{H_{prs}}$ changes the absolute scores but does not alter the overall conclusion. The large gap between frame-level and event-level performance persists, emphasizing that event-level VAD is not a trivial extension of frame-level thresholding and that explicit event-aware modeling is necessary for reliable anomaly detection in real-world settings.

\section{Conclusion}
\label{sec:conclusioin}

This work highlights the limitations of frame-level evaluation in pose-based Video Anomaly Detection (VAD), showing that high frame-level metrics often overestimate real-world effectiveness due to fragmented and temporally inconsistent detections. We propose an event-centric framework, including an event-based benchmark audit, a score-refinement pipeline, and a Dual-Branch event-aware model, to generate coherent event-level predictions. By adapting Temporal Action Localization metrics such as tIoU and multi-threshold $F_1$, we provide a realistic and operationally meaningful evaluation. Our results reveal a substantial gap between frame-level and event-level performance, emphasizing the need for event-focused methods that capture the temporal dynamics of anomalies for reliable real-world deployment.

\section*{Acknowledgment}
This research is supported by the National Science Foundation (NSF) under Award Number 2329816.
{
    \small
    \bibliographystyle{ieeenat_fullname}
    \bibliography{main}
}


\end{document}